\documentclass[10pt,twocolumn,letterpaper]{article}

\usepackage{iccv}
\usepackage{times}
\usepackage{epsfig}

\usepackage{booktabs}
\usepackage{multirow}

\usepackage{subcaption}

\usepackage{hyperref}
\usepackage{url}

\usepackage{tikz}
\usepackage{comment}
\usepackage{amsmath,amssymb} % define this before the line numbering.
\usepackage{color}

\usepackage{cuted}
\usepackage{capt-of}

\usepackage{breqn}

\usepackage{enumitem}
\usepackage{makecell}

\usepackage{blindtext}% for dummy text only
\usepackage{graphicx}
\usepackage{caption}

\iccvfinalcopy % *** Uncomment this line for the final submission

\begin{document}

%%%%%%%%% TITLE
\title{CLIP goes 3D: Leveraging Prompt Tuning for Language Grounded \\ 3D Recognition}

\author{Deepti Hegde$^*$, Jeya Maria Jose Valanarasu$^*$, Vishal M. Patel\\
Johns Hopkins University\\
{\tt\small {dhegde1,jvalana1,vpatel36}@jhu.edu}
}

\ificcvfinal\thispagestyle{empty}\fi

%%%%%%%%% ABSTRACT
\twocolumn[{%
\renewcommand\twocolumn[1][]{#1}%
\maketitle
\begin{center}
    \centering
    \captionsetup{type=figure}
    \vspace{-1 em}
    \includegraphics[width=1\textwidth, page=2]{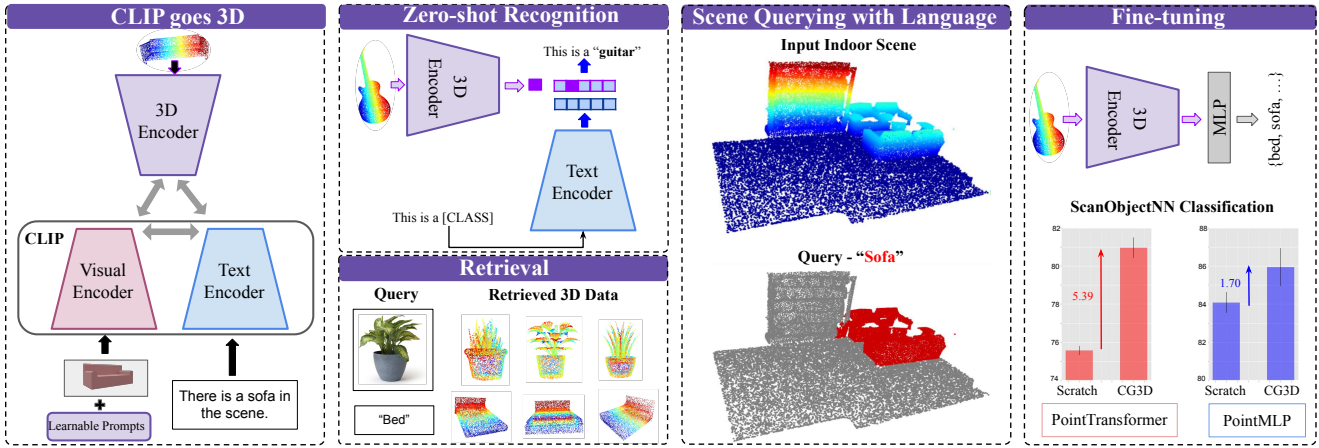}
    \captionof{figure}{Overview of our proposed framework CLIP goes 3D (CG3D). We introduce a 3D Encoder in the CLIP framework and pre-train it using natural language supervision while also leveraging CLIP's pre-trained visual encoder. CG3D solves various practical tasks like zero-shot 3D recognition, 3D point cloud retrieval, scene querying with natural language, Moreover, it can serve as a strong initial weight for standard fine-tuning tasks. }
\end{center}%
\label{teaser}

}]
\maketitle
\def\thefootnote{*}\footnotetext{Equal Contribution}\def\thefootnote{\arabic{footnote}}
\begin{abstract}
   Vision-Language models like CLIP have been widely adopted for various tasks due to their impressive zero-shot capabilities. However, CLIP is not suitable for extracting 3D geometric features as it was trained on only images and text by natural language supervision. 
   We work on addressing this limitation and propose a new framework termed CG3D (CLIP Goes 3D) where a 3D encoder is learned to exhibit zero-shot capabilities. CG3D is trained using triplets of pointclouds, corresponding rendered 2D images, and texts using natural language supervision.  To align the features in a multimodal embedding space, we utilize contrastive loss on 3D features obtained from the 3D encoder, as well as visual and text features extracted from CLIP.  We note that the natural images used to train CLIP and the rendered 2D images in CG3D have a distribution shift. Attempting to train the visual and text encoder to account for this shift results in catastrophic forgetting and a notable decrease in performance.  To solve this, we employ prompt tuning and introduce trainable parameters in the input space to shift CLIP towards the 3D pre-training dataset utilized in CG3D. We extensively test our pre-trained CG3D framework and demonstrate its impressive capabilities in zero-shot,  open scene understanding, and retrieval tasks.  Further, it also serves as strong starting weights for fine-tuning in downstream 3D recognition tasks. Codes and pre-trained models can be found here: \href{https://github.com/deeptibhegde/CLIP-goes-3D}{https://github.com/deeptibhegde/CLIP-goes-3D}.

\end{abstract}

%show that even weak, category level image-shape pairs and their captions are sufficient to transfer semantic understanding of objects from images to point clouds to

%%%%%%%%% BODY TEXT
\section{Introduction}
\label{sec:intro}

For many tasks in 2D vision, the most efficient and
accurate results are now obtained by adapting foundation
models \cite{ramesh2022hierarchical, ramesh2021zero, alayrac2022flamingo, yuan2021florence} which are pre-trained on large-scale data. There is currently a significant amount of research focused on efficiently adapting foundation models for specific 2D vision tasks \cite{zang2022unified, sohn2022visual, xing2022class, jia2022visual}, rather than developing new supervised methods from scratch. These approaches seeks to build on existing models and leverage their pre-trained features to achieve better performance on target tasks with less data and computation requirements. These recent trends in vision is actually similar to trends that were observed in natural language processing (NLP) a few years ago. In NLP, foundation models have been dominating since 2018 as models like BERT \cite{devlin2018bert} and GPT-3 \cite{brown2020language} showed exceptional ability to accomplish various NLP tasks, such as question answering, sentence prediction, sentiment classification, etc. Also, foundation models  pre-trained on multimodal data like images and text have been useful by exhibiting impressive zero-shot capabilities. In particular, Contrastive Language Image Pre-training (CLIP) \cite{radford2021learning} has been applied to various 2D tasks, including image classification \cite{zhou2022detecting}, object detection \cite{tevet2022motionclip, guzero}, image segmentation \cite{liang2022open, xie2022clims, zhou2021denseclip}, image retrieval \cite{lei2021less, luo2022clip4clip}, and visual question answering \cite{mokady2021clipcap, cho2022fine}.  These advances in foundation models in vision and NLP are yet to disrupt the field of 3D vision and understanding. In this work, we try to bridge this gap and focus on answering the following question: \textit{How can we build a 3D network that can possess similar functionalities as that of a foundation model like CLIP?}

3D visual understanding has many practical applications in robotics \cite{kastner20203d, zhou2021path}, augmented reality \cite{koniarski2022feature, wang2023pointshopar, you2022robot}, and autonomous driving \cite{qi2017pointnet, qi2018frustum, shi2019pointrcnn, milioto2019rangenet++}. Comprehending the semantics and characteristics of each point in a 3D space is crucial for addressing a wide range of issues in downstream tasks. Thus, there lies several use-cases for a potential 3D foundational model similar to foundation models for vision and NLP.  A powerful 3D network with zero-shot capabilities does not only help improve the performance of existing 3D backbones but also enables open 3D scene understanding and 3D retrieval tasks. However, the development of foundational models for 3D understanding faces several challenges, including the limited availability of 3D data compared to images. While CLIP was able to leverage the vast amount of images available on the internet to create a large pre-training dataset of image-caption pairs, a similar approach cannot be directly applied to pre-train a 3D encoder with texts due to the scarcity of 3D data. There have been some recent works like PointCLIP \cite{zhang2021pointclip} which tried to utilize CLIP's zero-shot capabilities for 3D zero-shot problems. PointCLIP directly uses the depth maps of a 3D point cloud on 2D visual encoder of CLIP. While it provides a quick and simple solution for 3D zero-shot problems, it lacks the characteristics of a foundational model since it cannot be used for 3D fine-tuning tasks or for 3D open scene understanding. Furthermore, it does not possess the ability to extract any 3D geometric features relevant for downstream tasks in 3D understanding.

To this end, we propose a new pre-training framework termed CG3D (CLIP Goes 3D) that trains a 3D encoder using natural language supervision while also leveraging CLIP's knowledge.  We begin by creating a pre-training dataset consisting of triplets of 3D point clouds, images, and corresponding text descriptions. We use point clouds from ShapeNet \cite{shapenet2015} as our 3D data and curate its corresponding rendered 2D image and a caption. Since ShapeNet consists of textured CAD models, we render random views of each object to use as the image pair. Despite their distinct properties, both a 3D point cloud and an image of the same object share common semantic attributes. This is affirmed by the success of tasks such as single image point cloud reconstruction \cite{mandikal20183d,fan2017point} as well as in the transfer of pre-trained weights from an image-based network to a 3D point cloud classification network, as seen in \cite{Xu2021Image2Point3P}. CG3D aims to ensure that there is similarity between 3D features and 2D features, as well as between 3D features and text features for objects of the same category, while being dissimilar for objects of different categories.  This contrastive approach to learning enables the 3D encoder to acquire zero-shot capabilities similar to those of CLIP.

The process of training the 3D encoder with contrastive loss and comparing 3D features to the 2D features from CLIP's visual encoder is a means of distilling CLIP's semantic features to the 3D encoder.  Although it would be efficient to train CLIP's visual encoder to align with the data distribution of 3D objects and their related images, we observed a significant decrease in performance when training both CLIP's visual and text encoders along with the 3D encoder in CG3D.  This can be explained as CLIP starts to catastrophically forget its previous features while being trained to shift to the new distribution. However, it is not ideal to keep the visual encoder completely frozen. Large-scale language models like CLIP are trained mostly on natural images, which differ in distribution to the graphically rendered views of 3D objects. To address this domain gap \cite{wang2018deep, VS_2022_WACV,vs2022target,VS_2021_CVPR}, we propose using prompt tuning techniques \cite{lester2021power, chen2022adaprompt} to shift the distribution in the input space before forwarding it to the visual encoder. We add visual prompts to the transformer backbone of CLIP's visual encoder thus adding only a small amount of parameters in the input space while keeping the weights of visual encoder of CLIP frozen. These parameters learn the shift in input distribution to suit CLIP so that the 3D pre-training is effective. To demonstrate the effectiveness of CG3D, we conduct several experiments. First, we show its zero-shot capabilities on synthetic and real object datasets like ModelNet \cite{wu20153d} and ScanObjectNN \cite{scanobjectnn}. Additionally, we showcase the 3D model's ability in open-scene comprehension by utilizing text-based queries, as well as its ability to conduct cross-modal 3D data retrieval while utilizing image or text queries. Further, the weights obtained from pre-training the 3D encoder using CG3D can also serve as effective initial weights when fine-tuning the model for other 3D tasks.

In summary, the following are our major contributions:

\begin{itemize}[topsep=0pt,noitemsep,leftmargin=*]

	\item We propose CG3D, a contrastive pre-training framework for training 3D networks using natural language supervision while also leveraging the knowledge of CLIP.

 \item We utilize prompt tuning to shift the input space of a pre-trained visual encoder from rendered images of CAD objects to natural images, allowing for more effective use of CLIP for 3D shapes.

 \item We conduct extensive experiments to demonstrate the versatile capabilities of CG3D. It exhibits strong zero-shot, 3D retrieval and 3D scene understanding capabilities with language. CG3D also acts as strong starting weights for multiple 3D recognition tasks.

\end{itemize}

\section{Related Works}

\noindent \textbf{Vision Language Models:}  The use of large-scale text-pre-training on attention-based models \cite{devlin2018bert,zhang2022opt} has led to the increasing popularity of vision-language models (VLM) due to their impressive performance in visual understanding tasks \cite{li2019visualbert,lu2019vilbert,radford2021learning}. Recent advancements in contrastive learning have enabled CLIP \cite{radford2021learning} to perform multimodal learning with 400M noisy data crawled from the web. CLIP has been extended for high efficiency model training and cycle consistency through various methods, such as ALBEF \cite{li2021align} and Cyclip \cite{goel2022cyclip}. BLIP \cite{li2022blip} includes text-to-image generation as an auxiliary task, which results in better performance by utilizing synthetic data as a bonus. Adopting VLM for 3D point cloud processing is still in its infancy. PointCLIP \cite{zhang2021pointclip} was the first method to adopt CLIP for 3D tasks. It directly uses the depth maps of 3D point clouds and uses it on the visual encoder of CLIP to perform zero-shot classification. Unlike this, we focus on using  a 3D encoder in the CLIP so that it can directly take in a 3D point cloud.

\noindent \textbf{3D Point cloud processing methods:}
In general, point cloud processing methods either  process the original point cloud sets directly \cite{qi2017pointnet, qi2017pointnet++} or transform the original point clouds into intermediate representations such as voxels \cite{maturana2015voxnet, shi2020pv} or images \cite{you2018pvnet, li2020end}. PointNet \cite{qi2017pointnet}  was a significant contribution to the field of point cloud processing, as it enabled the direct use of unordered point sets as input through shared MLPs. PointNet++ \cite{qi2017pointnet++} was later proposed as an extension of PointNet, incorporating a hierarchical feature learning approach that recursively captures local geometric structures. This feature representation method has proven to be effective due to its ability to capture multi-scale information, and it has been widely used in various point cloud applications \cite{wang2019dynamic, fan2021scf, xu2021paconv}. Recently, methods like PointTransformer \cite{zhao2021point} and PCT \cite{guo2021pct} have proposed transformer-based methods  showing a significant improvements in performance. The current state of the art method is PointMLP \cite{ma2022rethinking} which effectively uses a deep residual MLP network for point cloud analysis.
 
 There have also been pre-training methods which show 3D backbones can be pre-trained with unlabelled data to obtain strong initial weights to boost fine-tuning performance.  PointContrast \cite{xie2020pointcontrast} performs contrastive training \cite{chen2020simple} by pushing together heavily augmented views of the same sample and minimizing the similarity between views of other samples on point cloud scenes that have undergone rigid transformations. CrossPoint \cite{afham2022crosspoint} boosts point cloud classification performance by maximising the agreement between images and point cloud objects. PointBERT \cite{yu2022point} and PointMAE \cite{pang2022masked} leverage masked modelling methods to perform pre-training. Unlike these backbone or pre-training works, we focus on enabling zero-shot capabilities to given 3D encoder in our proposed CG3D framework.

\section{Method}

Our main objective is to train a 3D shape encoder to acquire shape characteristics that can effectively capture the geometric properties of point clouds while also aligning with CLIP's feature representation for each object category. In essence, we aspire to acquire features that are unique to each category yet unaffected by the mode of representation. To this end, we use point cloud-image-caption triplets to train the framework. Each element within a triplet of point cloud-image-caption is indicative of an object possessing specific semantic traits that are shared among the other objects in the collection. In this section, we first give an overview of the proposed CG3D framework, followed by details to effectively train the network. We then give details on the potential use-cases of CG3D.

\subsection{CG3D Framework}

\begin{figure}
    \centering
    \includegraphics[width=0.83\linewidth]{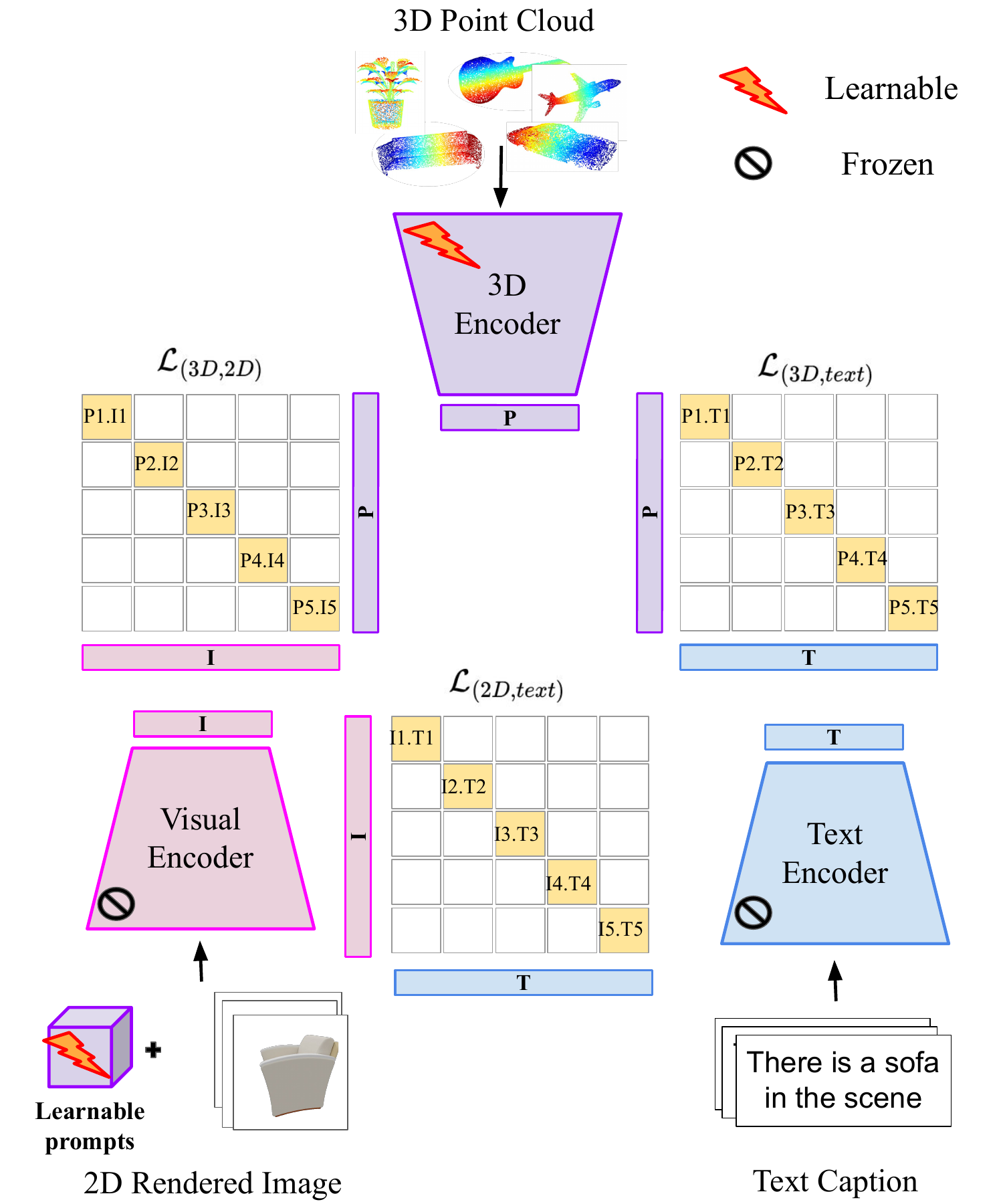}

    \caption{Overview of the proposed learning strategy in CG3D. Note that only the 3D Encoder and learnable visual prompts are trained while everything else is frozen.}
    \label{fig:mainfig}

\end{figure}

The CG3D framework as illustrated in Fig \ref{fig:mainfig} consists of 3 networks - 3D shape encoder, visual encoder and text encoder from CLIP.

\noindent \textbf{3D Encoder} takes in a 3D point cloud as the input. To capture the essential shape characteristics of an object, we employ a 3D encoder that is specifically designed to analyze point clouds and generate a feature vector representing the object. Our framework is agnostic with the choice of 3D encoder, and an added projection layer ensures the output feature dimension remains consistent. 

\noindent \textbf{2D Visual Encoder}  takes in the corresponding rendered image of the 3D pointcloud as the input. Although shape features are essential in representing point clouds, image backbones from vision-language models trained on large amounts of data offer powerful feature representations of images that are semantically correlated with text. We employ CLIP's visual encoder as it is robust and is pre-trained on a massive amount of data. By utilizing the visual encoder, we can acquire highly effective and implied representations of categories present in point cloud datasets, which are then use to align with the 3D features. Note that CLIP provides both ResNet and ViT weights for visual encoder.  

\noindent \textbf{Text Encoder} takes in the corresponding text caption of the 3D point cloud as input. Adopting natural language supervision for feature learning on images has been successful in training models to grasp visual concepts that can be depicted in both images and text \cite{radford2021learning}. In CLIP, the text encoder is trained to correspond text descriptions with images, which we use out-of-the-box under the same configuration as \cite{radford2021learning}.

\subsection{Training the 3D Encoder}

Our main training objective is to align the 3D point clouds with their corresponding category level images and texts. This alignment happens in a common embedding space to which data from each modality is projected from the modality-specific encoder and a projection head. Our training strategy relies on contrastive learning, which incentivizes cross-modal features of the same pair to be in close proximity to one another in the embedding space while keeping apart samples belonging to other pairs. We formulate the proposed losses below.
\setlength{\belowdisplayskip}{0pt} \setlength{\belowdisplayshortskip}{0pt}
\setlength{\abovedisplayskip}{0pt} \setlength{\abovedisplayshortskip}{0pt}
Consider a set of $N$ pointcloud-image-text triplets $\{x_i^{3D},x_i^{2D},x_i^{text}\}_{i=1}^{N}$, where $x_i^{3D}$ represents a pointcloud, $x_i^{2D}$ is the corresponding rendered image, and $x_i^{text}$ is the corresponding text. Let the encoder for each modality be represented as $\phi_{3D}$, $\phi_{2D}$, and $\phi_{text}$. We obtain the feature representation of each sample in a common embedding space by projecting the encoded feature to a common dimension represented by: 
\begin{equation}
    f_i^{3D} = \psi_{3D}(\phi_{3D}(x_i^{3D})) 
\end{equation}
\begin{equation}
    f_i^{2D} = \psi_{2D}(\phi_{2D}(x_i^{2D})) 
\end{equation}
\begin{equation}
    f_i^{text} = \psi_{text}(\phi_{text}(x_i^{text})) 
\end{equation}

where  $i$ ranges from 1 to the number of samples $N$ and $\psi$ is the projection operation for each modality. Through normalization, we constrain the output of each projection network to reside within a unit hypersphere, enabling us to measure feature similarity using the inner product. The contrastive losses (InfoNCE objective \cite{oord2018representation}) between 3D-image features; and 3D-text features are defined by:
\begin{equation}
    \begin{aligned}
    \mathcal{L}_{(3D,2D)} = \frac{1}{2} \sum_{(f^{3D}, f^{2D}) \in B}  \text{NCE} (f^{3D},f^{2D}) + \\  \text{NCE} (f^{2D}, f^{3D}) 
    \end{aligned}
\end{equation}
\begin{equation}
    \begin{aligned}
    \mathcal{L}_{(3D,text)} = \frac{1}{2} \sum_{(f^{3D}, f^{text}) \in B}  \text{NCE} (f^{3D},f^{text}) + \\  \text{NCE} (f^{text}, f^{3D}) 
    \end{aligned}
\end{equation}

    where $f^{3D}$, $f^{2D}$, and $f^{text}$ are the projected 3D, image and text features respectively. $B$ corresponds to the batch. NCE loss is defined as:
\begin{equation}
    \text{NCE} (f^{A}, f^{B}) = -  \log \frac{exp(\langle f^{A}, f^B_+ \rangle/\tau)}{\sum_{f \in (f^B_+, f^B_-) }exp( \langle f^{A},f \rangle /\tau)}
\end{equation}
  where $A$, $B$ are two different modalities and   $f^{A}$, $f^{B}$ are their corresponding features. $\tau$ is the temperature hyper-parameter, $f^B_+$ are the positive embeddings from modality $B$ overlapping with modality $A$, and $f^B_-$ are the negatively embeddings formed while pairing with modality $A$. For example, in $\mathcal{L}_{(3D,2D)}$ the positive pairs between 3D and 2D are formed by matching the features corresponding to same class while the rest are termed as negative pairs. The total loss $\mathcal{L}_{3D}$ used to train the 3D encoder is defined by:
  \begin{equation}
      \mathcal{L}_{3D} =   \mathcal{L}_{(3D,2D)} +   \mathcal{L}_{(3D,text)}
  \end{equation}

\subsection{Prompt Tuning for Visual Encoder}

The visual encoder in the CG3D framework takes in the rendered image of the 3D point cloud as its input.
While the CLIP visual encoder has been trained on vast amounts of internet data and is highly resilient, during pre-training of CG3D, it only deals with rendered images. As a result, fine-tuning the CLIP visual encoder to handle rendered images could improve the training process for the 3D encoder. One possible approach is to train the visual encoder by optimizing its weights using the CLIP loss function that computes the similarity between image and text features. However, when we tried this method, we noticed a substantial decrease in performance. This phenomenon can be explained by the fact that training the visual encoder of CLIP causes catastrophic forgetting. That is, the encoder loses all of its prior knowledge while attempting to adapt to the new data distribution. Typically, this issue can be avoided by increasing the amount of new data available for fine-tuning. However, it's not feasible to obtain a large enough 3D dataset that matches the scale of the massive image-text data used to train CLIP. Therefore, we concentrate on developing methods to effectively fine-tune the model with new pre-training data while keeping the visual encoder frozen.

Visual prompt tuning, as described in \cite{jia2022visual}, is a method that involves adding a small number of trainable parameters in the input space to fine-tune a base model for a specific task. In our proposed method for pre-training CG3D, we adopt this approach and modify the input space to better align with the visual encoder of the original CLIP model. This, in turn, allows the visual encoder to produce higher-quality features, which can enhance the training of the 3D encoder. We use deep prompting where we introduce learnable prompts as learnable tokens at every layer in the transformer layer in ViT (visual encoder). For an $i^{th}$ transformer layer $L_i$, we define the collection of learnable tokens at that layer as $P_i = \{ p_i^k , 1 \le k \le n\}$ where $p$ corresponds to individual tokens and $n$ is the total number of learnable tokens. The deep prompted visual encoder at $i^{th}$ can be represented as :
\begin{equation}
    [y_i] = L_i([y_{i-1}, P_{i-1}])
\end{equation}
where $y_i$ is the output and $y_{i-1}$ is the input to the current layer. Prompt tokens $P$ are trained along with the 3D encoder in our CG3D framework. We use the original CLIP loss which is a contrastive loss between the image and text features to train these prompts. We formulate the loss used to train the prompts $\mathcal{L}_{P}$ as:
\begin{equation}
    \begin{aligned}
    \mathcal{L}_{P} = - \sum_{(f^{2D}, f^{text}) \in B} (\log \text{NCE} (f^{text},f^{2D}) + \\ \log \text{NCE} (f^{2D}, f^{text}) )
    \end{aligned}
\end{equation}
 
where  where  $f^{2D}$ and $f^{text}$ are the image and text features respectively. $B$ corresponds to the batch.

\subsection{Using CG3D }

\subsubsection{Zero-shot 3D Recognition}

Zero-shot 3D classification refers to the method of classifying 3D objects without requiring any previous training on those specific objects. It has a number of use-cases in robotics and autonomous systems where objects need to be recognized quickly and accurately, without the need for extensive training on new objects. CG3D enables zero-shot 3D recognition directly with a 3D encoder which extracts shape features from the input point clouds. For zero-shot inference, we only use the 3D and text encoder from our proposed CG3D framework. The model takes as input a point cloud test sample, denoted as $x_{test}$, and a set of prompts $\textbf{T} = \{t_i, 1 \le i \le N\} $ where $t_i$ represents individual text prompts and $N$ is the total number of text prompts. Each prompt $t_i$ is denoted in the format of "This is a {OBJECT}", where {OBJECT} represents the name of a test class. The model's inference procedure is similar to that of CLIP, where it calculates the similarity score between each prompt and the test sample and selects the prompt that yields the highest score as the final prediction. This is formulated as:
\begin{equation}
    y_{pred} = \text{max}(\text{softmax}(\langle f^{3D} , \textbf{F}^{text} \rangle))
\end{equation}
where $f^{3D}$ and $\textbf{F}^{text}$ are the feature vectors of the point cloud and text inputs respectively, $y_{pred}$ is the class prediction. Note that  $\textbf{F}^{text}$ is actually a collection of feature vectors collected from forwarding the text queries $T$ to the text encoder. This process has been summarized in Fig 1. 

\subsubsection{Scene Querying with Language}

\begin{figure}[htbp]
    \centering
    \includegraphics[width=0.8\linewidth]{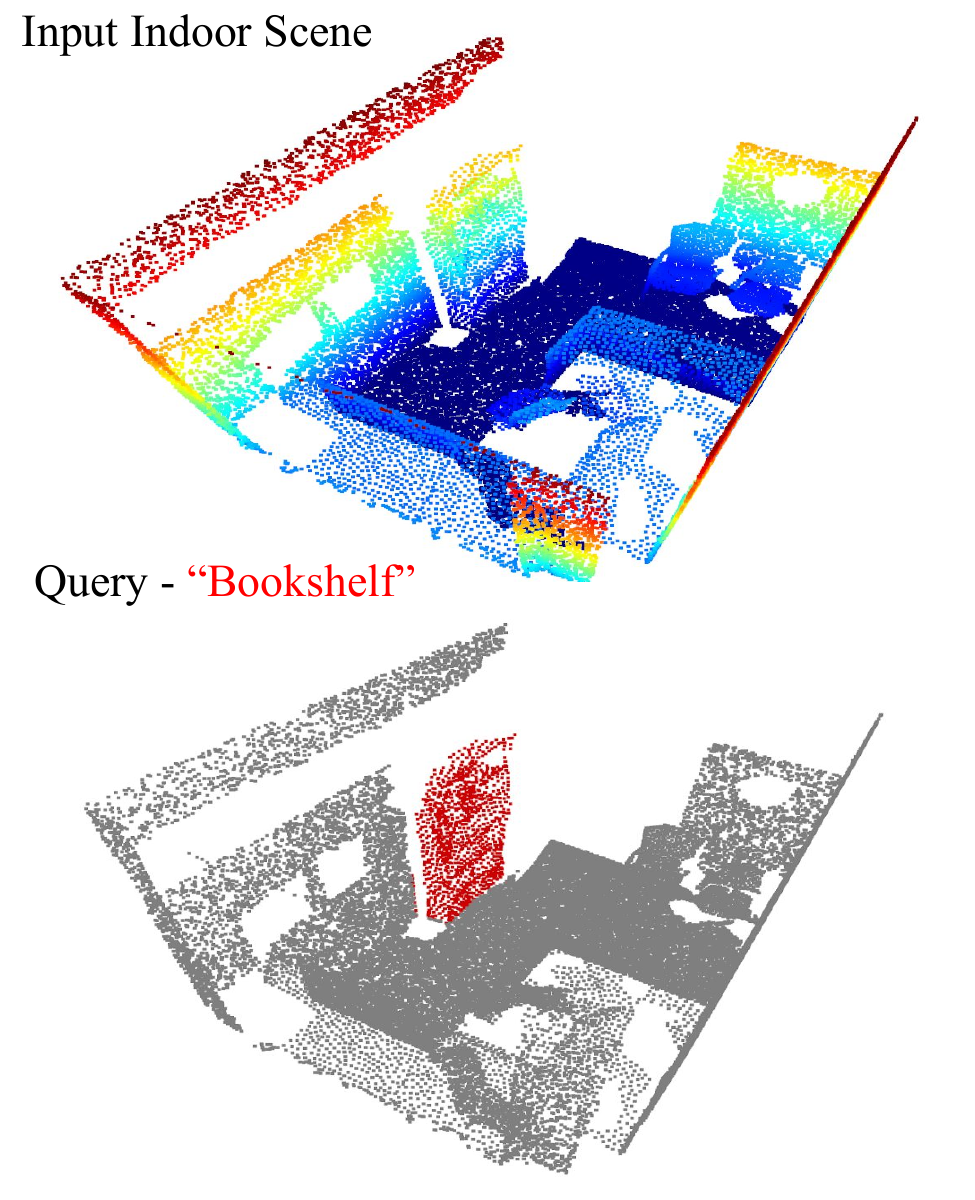}
    \caption{Example of scene querying with text for a random indoor scene from S3DIS \cite{2017arXiv170201105A} dataset. }
    \label{open}
\end{figure}

3D scene understanding is an important task in computer vision. It is crucial for enabling human-robot interaction and facilitating intuitive human-machine interfaces. In particular, querying a scene with language queries to understand the key details of a scene is a practically useful task. One aspect of this is answering queries such as "What is the location of the sofa?" or "Where can the chair be found?", which can help an individual in comprehending the environment or allow a robot to interact with it intelligently. By accurately identifying objects and their locations within a scene, robots can perform a wide range of tasks such as manipulation, navigation, and object recognition. Meanwhile, individuals can use interactive query and visualization tools to better understand and analyze complex 3D scenes.

CG3D enables zero-shot scene understanding with language queries. Without training the model on indoor scenes or using direct supervision, we show that a pre-trained CG3D framework can be used to understand scenes with text queries. Given an input scene, first we use k-means clustering to divide the scene into meaningful segments. Next, we feed forward all these clusters to the 3D encoder in our CG3D framework and get a set of 3D features $\textbf{F}^{3D} = \{ f^{3D}_i, 1 \le i \le k \}$. Here, k is the total number of clusters obtained from the scene and $f^{3D}_i$ is the 3D feature vector from 3D encoder of CG3D for the $i^{th}$ cluster. We also pass the text query to the text encoder to obtain $f^{text}$. Now, we match these 3D features with the text feature we obtain by forwarding the input query to the text encoder which can be denoted as:
\begin{equation}
    y_{pred} = \text{max}(\text{softmax}(\langle f^{text} , \textbf{F}^{3D} \rangle))
\end{equation}

where $y_{pred}$ is the predicted class. To demonstrate an example for scene understanding with language using CG3D, we pick a random indoor scene as seen in Fig \ref{open} from the S3DIS \cite{2017arXiv170201105A} dataset and use the query "Bookshelf". It can be observed that the model correctly classifies which cluster is the bookshelf in the scene. More analysis can be found in the supplementary material. Although CG3D is not explicitly designed to provide a complete solution for open world 3D scene understanding, it facilitates scene querying using natural language.

\subsubsection{Retrieval}

3D point cloud retrieval is the process of searching and retrieving 3D point cloud data that is similar to a given query. The query here can be of different modality like an image or a text. This has several practical applications like matching a real-world object with its corresponding 3D point cloud in a virtual environment to help create more realistic and accurate augmented and virtual reality experiences. 

To retrieve data using CG3D, we utilize the pre-trained encoders to obtain feature representations for both the query and the 3D point clouds. Specifically, we feed the image or text query into the corresponding encoder to obtain its feature vector, and we do the same for the 3D point clouds using the 3D encoder. The point clouds being forwarded to the 3D encoder constitute the complete database from which we are retrieving the relevant shapes. Next, we obtain the similarity score between these query feature and the 3D features and select the point clouds with the highest similarity scores as the output. 

\begin{figure}
    \centering
    \includegraphics[width=\linewidth]{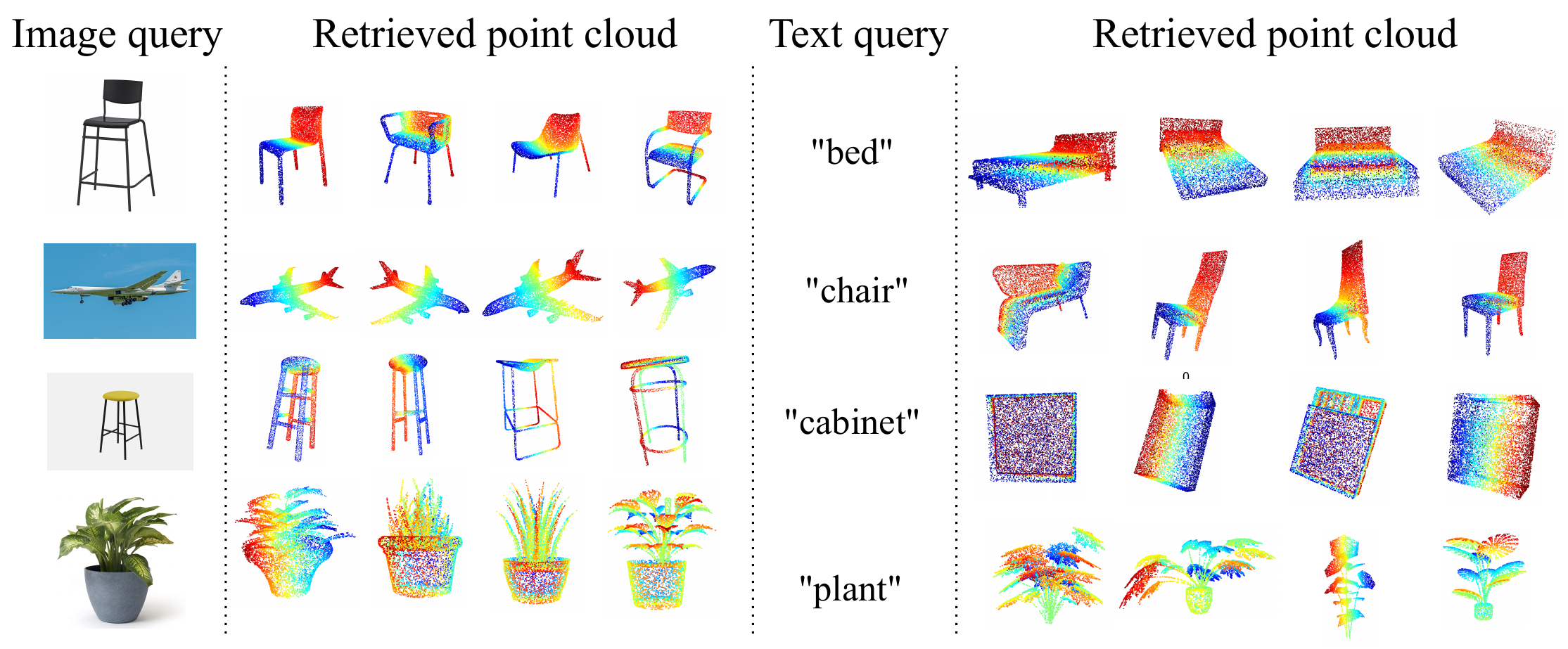}
    \caption{Retrieving point clouds from a 3D database (ModelNet40) using random image and text queries.}
    \label{ret}
    \vspace{-1 em}
\end{figure}

 In Fig \ref{ret}, we demonstrate the effectiveness of CG3D in retrieving relevant 3D point clouds. We randomly select some images from the internet as query images, and ModelNet40 serves as our 3D database for retrieval. We pick the top four 3D points that are of the highest similarity to the input query and display them in Fig \ref{ret}. Similarly, we also use some random text queries and display the best matches. It can be observed that all the retrieved point clouds are of very high similarity to the input query proving the effectiveness of CG3D.

\subsubsection{Fine-tuning for Supervised Tasks}

Pre-training techniques are an effective strategy for enhancing the performance of fine-tuning in 3D computer vision tasks. Pre-training models on large datasets of unlabeled images can help them learn generic and transferable features, making them more robust to variations in data and enabling them to generalize well to new tasks and datasets. 

Although the main objective of CG3D is  its zero-shot capabilities,  it also has the potential to serve as a valuable starting point for fine-tuning 3D models for downstream tasks. This is due to the excellent feature representation capabilities of the 3D encoder, which has been pre-trained using natural language supervision in the CG3D framework. Additionally, CG3D is model-agnostic, meaning that any 3D backbone can be pre-trained using CG3D, and the resulting weights can be used as a starting point for downstream tasks. We present multiple experiments that demonstrate the effectiveness of using CG3D for fine-tuning tasks.

% \subsubsection{Training the shape encoder}
% \subsubsection{Text supervision}
% \subsection{Tuning CLIP's visual encoder}

% \begin{figure}
%     \centering
%     \includegraphics[width=\linewidth]{figures/zeroshot.pdf}
%     \caption{Zero shot}
%     \label{fig:zero}
% \end{figure}

\section{Experiments and Results}

\subsection{Datasets}
\noindent\textbf{Pre-training dataset:}  We choose ShapeNet \cite{shapenet2015} as the pre-training dataset due to its large number of classes and samples. ShapeNet consists of textured CAD models of  55 object categories and 52,460 total samples. We sample point clouds of a fixed size from each object mesh and normalize them to fit into a unit sphere. We render the colored CAD model views in Blender to obtain a image pair for each point cloud, following \cite{gao2022get3d}. The text captions of each point cloud-image pair are framed as a descriptive sentences obtained from a set of standard templates such as ``A photo of a \{OBJECT\}". Each input point cloud is augmented with standard techniques such as object scaling, rotation, random drop, and perturbations.

\noindent \textbf{Fine-tuning datasets:} We perform zero shot (ZS) classification  and downstream fine-tuning on the popular 3D datastes ModelNet40 \cite{wu20153d} and ScanObjectNN \cite{scanobjectnn}. As is standard, we evaluate on the full dataset (ModelNet40) as well as a 10-class subset (ModelNet10) for ZS classification. ModelNet40 consists of 12,311 synthetic meshes of common objects from 40 categories. Each mesh is downsampled and normalized to fit a unit sphere. ScanObjectNN is a real-world point cloud dataset of objects from laser-scanned indoor scenes. 

\subsection{Implementation Details}
There exist several vision-language models that can be considered variants of CLIP that give superior zero-shot performance on images \cite{li2021align,mu2022slip,declip}. We leverage the pre-trained visual and text encoder weights from SLIP \cite{mu2022slip} to train CG3D, due to its performance and flexibility. We specifically choose ViT-Base \cite{dosovitskiy2020image} as the image backbone.%,and demonstrate pre-training and fine-tuning for 3D recognition on two recent point cloud classification networks PointTransformer \cite{zhao2021point} and PointMLP \cite{ma2022rethinking}.

 \noindent\textbf{Pre-training: }  During pre-training, the visual prompt parameters and the parameters of the 3D encoder are tuned under different optimizers in alternate iterations. This is due to the fact that they are each supervised by disjoint loss functions and require different learning rates. We append 5 learnable prompt tokens at the input of every encoder layer in ViT, and initialize them randomly. The visual prompts are tuned using the SGD \cite{ruder2016overview} optimizer under a cosine annealing \cite{loshchilov2016sgdr} scheduler, with learning rate of $2\times 10^{-3}$, weight decay $10^{-4}$ and an minimum learning rate of $10^{-6}$. We follow the training convention of each 3D backbone for training the individual backbone. In  case of PointTransformer \cite{zhao2021point} the network is tuned using the AdmaW \cite{loshchilov2017decoupled} optimizer under a cosine  annealing \cite{loshchilov2016sgdr} scheduler with a learning rate of $5\times 10^{-5}$ and a weight decay of $0.05$, with a minimum learning rate of $10^{-6}$. The PointMLP backbone is tuned under the same otimizer-scheduler scheme with learning rate $10^{-4}$ and weight decay $0.01$.  The entire framework is pre-trained for 100 epochs with a batch size of $32$.

\noindent\textbf{Fine-tuning: } PointMLP is finetuned using the SGD optimizer and cosine scheduler with a learning rate of $0.02$, a weight decay of $2\times 10^{-4}$, and a minimum learning rate of $5\times 10^{-3}$. PointTransformer is fine-tuned using the AdamW optimizer and cosine scheduler, with learning rate $2\times 10^{-4}$, a weight decay of $0.05$ and a minimum learning rate of $10^{-6}$. Each network is fine-tuned for 300 epochs with a batch size of 32 for PointMLP and 64 for PointTransformer. Our method is prototyped in PyTorch and all our experiments are performed in a 8 GPU NVIDIA A100 cluster. 

\subsection{Zero-shot Experiments}

We present the results of zero-shot experiments conducted on test distributions of ModelNet10, ModelNet40, and ScanObjectNN in Table \ref{zs}. We experiment with two backbones: PointTransformer and PointMLP pre-trained with CG3D. Note that the previous method PointCLIP uses a 2D depth map and CLIP's visual encoder to get the prediction. We directly use the 3D encoder and extract relevant 3D shape features to perform the zero-shot classification. This gives a significant improvement over PointCLIP with an increase of $37.1 \%$ on ModelNet10, $30.4 \%$ on ModelNet40, and  $10.2 \%$ on ScanObjectNN.

\begin{table}[]

\centering

\resizebox{\columnwidth}{!}{%
\begin{tabular}{c|ccc}
\midrule[1pt]\toprule[0.1pt]
\multirow{2}{*}{Method} & \multicolumn{3}{c}{Zero-shot performance}                                                                      \\ 

\cline{2-4} 
                        & \multicolumn{1}{c|}{MN10} & \multicolumn{1}{c|}{MN40} & \multicolumn{1}{c}{ScanObjectNN}  \\ \hline

PointCLIP \cite{zhang2021pointclip}              & \multicolumn{1}{c|}{30.2}      & \multicolumn{1}{c|}{20.2}      & \multicolumn{1}{c}{15.4}                 \\ 
PointTransformer \cite{zhao2021point}+ CG3D           & \multicolumn{1}{c|}{67.3}       & \multicolumn{1}{c|}{50.6}      & \multicolumn{1}{c}{25.6}
\\ 
PointMLP \cite{ma2022rethinking} + CG3D           & \multicolumn{1}{c|}{64.1}       & \multicolumn{1}{c|}{50.4}      & \multicolumn{1}{c}{25.0} \\ \midrule[0.1pt]\toprule[1pt]
\end{tabular}
}

\caption{Comparison of zero-shot classification performance of CG3D against that of PointCLIP for the ModelNet10, ModelNet40, and ScanObjectNN datasets.}
\label{zs}
\vspace{-0.5 em}
\end{table}

\subsection{Fine-tuning Experiments}

\begin{table}[]
\resizebox{\columnwidth}{!}{%
\begin{tabular}{ccc}
\midrule[1pt]\toprule[0.1pt]
\multirow{2}{*}{Method} & \multicolumn{2}{c}{Overall accuracy} \\ \cline{2-3} 
                        & ModelNet40          & ScanObjectNN         \\ \hline
                        Pointnet  \cite{qi2017pointnet}            & 89.2                     &   68.0 \\  
Pointnet$++$  \cite{qi2017pointnet++}            &   90.5                  &   77.9                   \\
PointCNN  \cite{li2018pointcnn}             &  92.2                   &  78.5                    \\
DGCNN \cite{wang2019dynamic}              &  92.9                   &    78.1                  \\
Point-BERT  \cite{yu2022point}             & 93.2                    &  83.07                    \\
Point-MAE \cite{pang2022masked}              & 93.8                    &  85.18                    \\
\hline
PointTransformer \cite{zhao2021point}        &  91.62 $\pm$ 0.29                    &   75.56 $\pm$ 0.24                   \\
PointTransformer \cite{zhao2021point} $+$ CG3D                &  92.93 $\pm$ 0.06                   &   80.95 $\pm$ 0.54                   \\ \hline
PointMLP \cite{ma2022rethinking} &   92.61 $\pm$ 0.13                  & 84.08 $\pm$ 0.55                     \\
PointMLP \cite{ma2022rethinking} $+$ CG3D         &  93.35 $\pm$ 0.18                   &  85.78 $\pm$ 0.75                    \\ \midrule[0.1pt]\toprule[1pt]
\end{tabular}
}

\caption{Comparison of fine-tuning performance of CG3D with initial weights on ModelNet40 and ScanObjectNN (hardest variation: PB-T50-RS) against previous methods.}
\label{fine}
\vspace{-1 em}
\end{table}

We present the results of our fine-tuning experiments in Table \ref{fine} on both synthetic (ModelNet40) and real (ScanObjectNN)  datasets. For ScanObjectNN, we pick the hardest variant PB-T50-RS for our experiments. We compare against leading backbones as well as pre-training methods like Point-BERT and Point-MAE. We show fine-tuning performance on two backbones PointTransformer and PointMLP. It should be noted that our framework was not primarily developed to be a  pre-training strategy, but rather to enable zero-shot capabilities for a 3D encoder. Even then, our framework demonstrates competitive performance as a pre-training strategy as observed in Table \ref{fine}. In particular, we obtain a boost of $5.39 \%$ , $1.31 \%$ while using CG3D starting weights than random weights for PointTransformer on ScanObjectNN and ModelNet40 respectively. Also, we obtain a boost of $1.7 \%$ , $0.74 \%$ while using CG3D starting weights than random weights for PointMLP on ScanObjectNN and ModelNet40 respectively. These observations show the versatility of our proposed framework. We re-run the from-scratch expriments for the PointTransformer and PointMLP networks. To account for variability,  we conducted the finetuning experiments for PointTransformer and PointMLP three times each, starting from scratch and also starting with CG3D weights. To account for this, we have reported the mean and standard deviation of the results obtained from these experiments.

% \begin{table}[]
% \resizebox{\columnwidth}{!}{%
% \begin{tabular}{ccc}
% \hline
% \multirow{2}{*}{Method} & \multicolumn{2}{c}{16 shot accuracy} \\ \cline{2-3} 
%                         & ModelNet40       & ScanObjectNN      \\ \hline
% PointTransformer        &                  &                   \\
% PointMLP                &                  &                   \\
% PointCLIP               &                  &                   \\\hline
% PointTransformer + CG3D &                  &                   \\
% PointMLP + CG3D         &                  &                  \\ 
% \end{tabular}
% }
% \end{table}

% \usepackage{multirow}

% \begin{figure*}
%     \centering
%     \includegraphics[width=\linewidth]{figures/scene_ret_main_cluster.drawio.pdf}
%     \caption{Caption}
%     \label{fig:my_label}
% \end{figure*}

\section{Discussion}

% Before contrastive training, the point cloud encoder is unable to produce discriminative features for all classes as seen in (a). After self-supervised contrastive training on ShapeNet, (b) the point cloud encoder produces features of samples from ModelNet40 with well-pronounced class separability.

\noindent \textbf{Ablation Study:}
We conduct an ablation study to analyze the role that each component of CG3D has on zero-shot performance. In Table \ref{tab:ablation}, we report the overall zero-shot accuracy  of the PointTransformer 3D encoder pre-trained under different loss configurations for the PB-T50-RS variant of ScanObjectNN. We start with the configuration of training the 3D encoder with 
$\mathcal{L}_{(3D,2D)}$ and $\mathcal{L}_{(3D,text)}$ individually. We note that training with $\mathcal{L}_{(3D,text)}$ gives a slight improvement over just training with $\mathcal{L}_{(3D,2D)}$. Next, we pre-train CG3D with both these losses which obtains better performance over individual configurations. After this, we incorporated visual prompts into the CG3D model and trained it using $\mathcal{L}_{P}$. This further improved the model's performance.

% We also compare performance with and without visual prompt tuning, which allows us to optimize the $\mathcal{L}_{(2D,text)}$ loss. It is clearly important to pre-train the network using both $\mathcal{L}_{(3D,text)}$ and $\mathcal{L}_{3D,image}$ loss, with the use of visual prompt tuning providing a significant improvement.

\begin{table}[htbp]
\centering
\resizebox{0.6\columnwidth}{!}{%
\begin{tabular}{ccccc}
\hline
 $\mathcal{L}_{(3D,2D)}$ &  $\mathcal{L}_{(3D,text)}$  & visual prompt & ZS  \\ \hline
           \checkmark &       $\times$                      & $ \times$ &   19.1      \\
           $\times$ &       \checkmark                  &  $\times$   &     19.7  \\
           
        \checkmark    &      \checkmark                         &  $\times$  &     23.9     \\
        
        \checkmark &       \checkmark                    &  \checkmark   &    25.6   \\\hline
\end{tabular}
}%
\vspace{-1 em}
\caption{Ablation Study on ScanObjectNN.}
\label{tab:ablation}
  \vspace{-1 em}
\end{table}

\begin{figure}[htbp]
    \centering
      \vspace{-2.5 em}
    \includegraphics[width=1\linewidth]{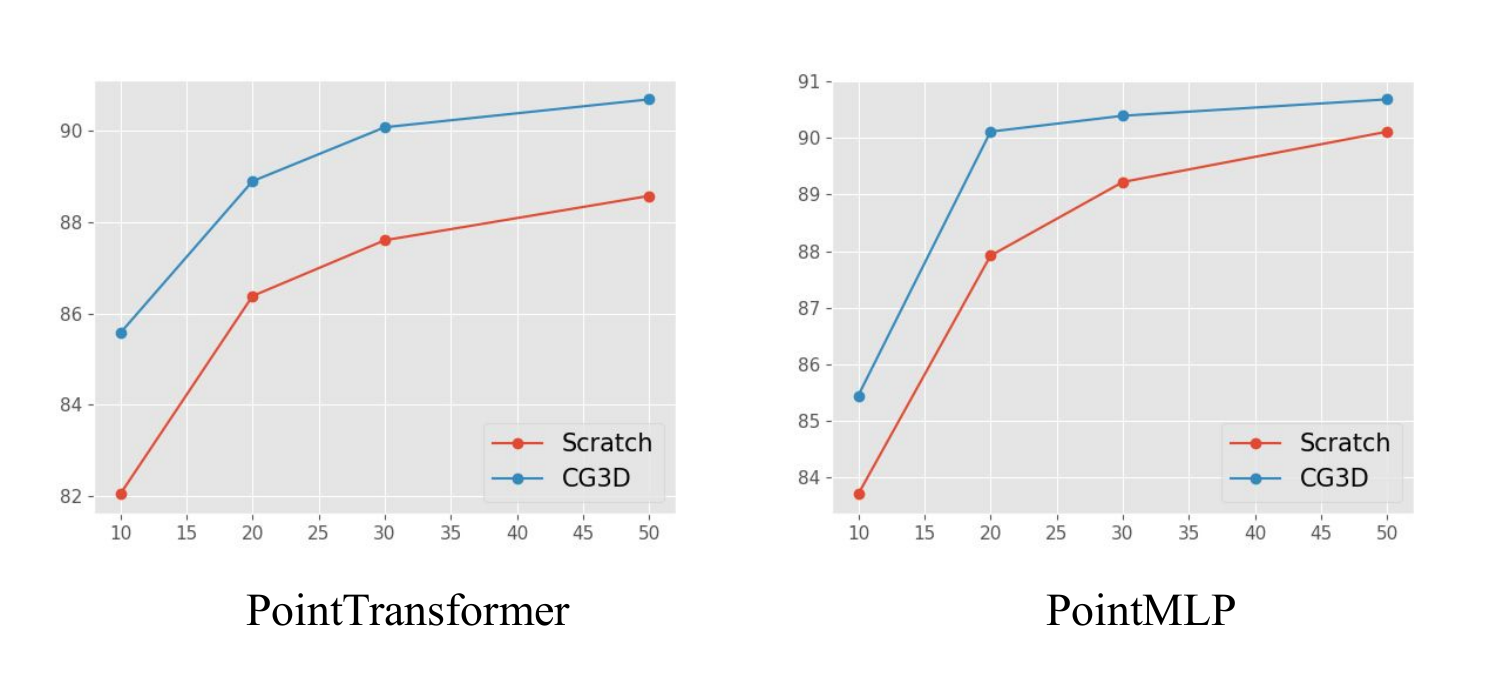}
    \vspace{-2.5 em}
    \caption{Experiments on data scarce setups on ModelNet40 with PointTransformer and PointMLP backbones. }
    \label{datascar}
\end{figure}

\noindent \textbf{Analysis on Less Data:} Pre-trained models are particularly useful when dealing with tasks that have limited access to data. To prove the effectiveness of CG3D in such scenarios, we conduct experiments on ModelNet40 with both PointMLP and PointTransformer as backbones.
We fine-tune each 3D backbone on different sub-sets $10 \%, 20 \%, 30 \%, 50 \%$ of the data and present the results in Fig. \ref{datascar}. It can be observed that the model trained with starting weights of CG3D always obtains better performance than starting with random weights across all configurations.

% \noindent \textbf{Analysis on Prompt Tuning}

\begin{figure}[htbp]
     \centering
     % \begin{subfigure}[b]{0.28\textwidth}
     %     \centering
     %     \includegraphics[width=\textwidth]{cvpr2023-author_kit-v1_1-1/latex/figures/umap_image.drawio.pdf}
     %     \caption{UMAP visualization of image features from CLIP's \textbf{pre-trained, frozen} ResNet50 visual encoder. }
     %     \label{fig:umap1}
     % \end{subfigure}
     % \hfill
     \begin{subfigure}[b]{0.4\linewidth}
         \centering
         \includegraphics[width=\textwidth]{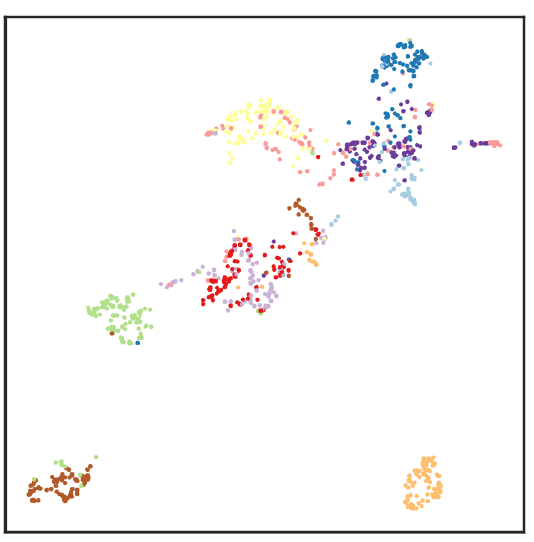}
         \caption{\textbf{Before} CG3D.}
         \label{fig:umap_point_1}
     \end{subfigure}
     \hfill
     \begin{subfigure}[b]{0.4\linewidth}
         \centering
         \includegraphics[width=\textwidth]{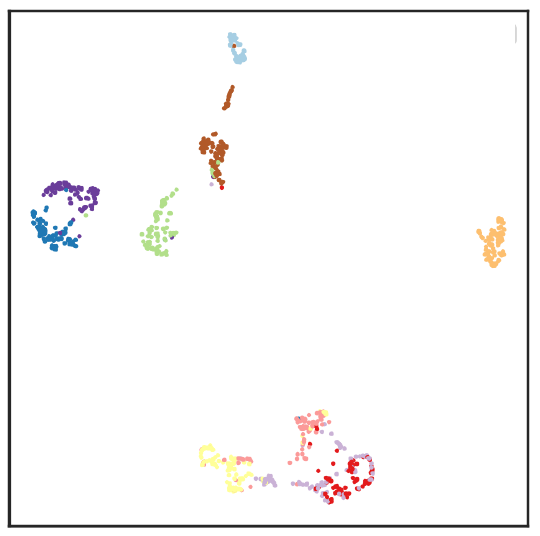}
         \caption{\textbf{After} CG3D.}
         \label{fig:umap_point_2}
     \end{subfigure}
     \vspace{-1 em}
        \caption{Comparison of UMAP embeddings of point cloud features from 3D encoder of CG3D while using ModelNet10 3D point clouds.  }
        \label{fig:umap_point}
        \vspace{-1 em}
\end{figure}

\noindent \textbf{Visualization of 3D Feature Representations:} For analysis of the quality of features learned by our method, we visualize the UMAP \cite{mcinnes2018umap} embeddings of 3D and image features extracted by CG3D while using ModelNet10. In Fig. \ref{fig:umap_point}, we plot the 3D feature learned by PointTransformer before and after pre-training with CG3D. Most features lack class separability before pre-training as can be seen in Fig. \ref{fig:umap_point_1}. After pre-training, the 3D encoder is able to produce class discriminative features even for unseen categories in ModelNet10 as seen in Fig. \ref{fig:umap_point_2}. 

\begin{figure}[htbp]
     \centering
     % \begin{subfigure}[b]{0.28\textwidth}
     %     \centering
     %     \includegraphics[width=\textwidth]{cvpr2023-author_kit-v1_1-1/latex/figures/umap_image.drawio.pdf}
     %     \caption{UMAP visualization of image features from CLIP's \textbf{pre-trained, frozen} ResNet50 visual encoder. }
     %     \label{fig:umap1}
     % \end{subfigure}
     % \hfill
     \begin{subfigure}[b]{0.4\linewidth}
         \centering
         \includegraphics[width=\textwidth]{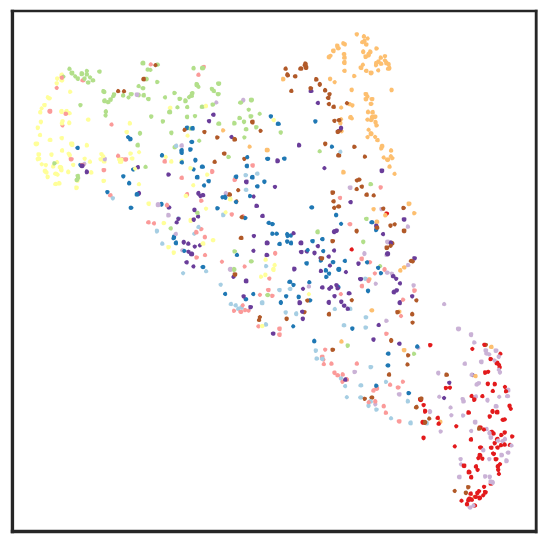}
         \caption{\textbf{Without} VPT.}
         \label{fig:umap_im_1}
     \end{subfigure}
     \hfill
     \begin{subfigure}[b]{0.4\linewidth}
         \centering
         \includegraphics[width=\textwidth]{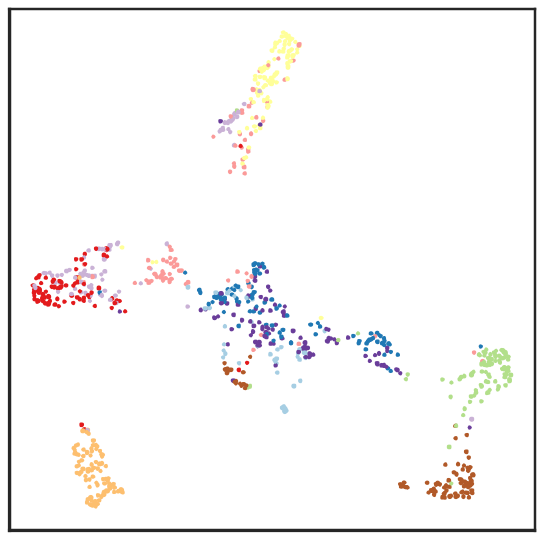}
         \caption{\textbf{After} contrastive VPT.}
         \label{fig:umap_im_2}
     \end{subfigure}
     \vspace{-1 em}
        \caption{Comparison of UMAP embeddings of visual encoder features of CG3D with ModelNet10 depth maps.}
        \label{fig:umap_im}
        \vspace{-1 em}
\end{figure}

\noindent \textbf{Effect of Prompt Tuning:} We visualize the image features learned by the visual encoder with and without the learned prompts in CG3D. Since textured CAD models are not available for all samples, we consider depth maps of the points projected in the 2D plane. Figure \ref{fig:umap_im_1} shows the depth map image features learned by the CLIP visual encoder. Since these images are visually dissimilar from natural images, the encoder fails to produce discriminative features. However, as seen in Figure \ref{fig:umap_im_2}, the visual encoder trained with visual prompts after CG3D pre-training produces features with improved class separability proving the effectiveness prompt tuning. 

% (a) with and (without performing prompt tuning on the visual encoder. (a)The out-of-the-box visual encoder fails to discriminate between depth images of different categories while (b) the visual encoder tuned with CG3D succeeds

% \noindent \textbf{Prompt Tuning with depth maps}

% \noindent \textbf{Concurrent Works:} We note that this topic is an active area of research and there are multiple concurrent works like ULIP, ConceptFusion, and OpenScene which also work towards using CLIP for 3D tasks. Although these papers are un-published at the time this paper is written, we highlight our differences with them in the supplementary material.

\noindent \textbf{Limitations:} We note that our pre-training dataset is still small in size, and consists of only simulated point cloud objects, thus limiting the potential of CG3D. To build a powerful foundation model for 3D, we need to work on data curation of
3D pointclouds, with corresponding images and text captions. We also focused on pre-training on objects and not scenes. Pre-training on scenes could open up interesting full scene understanding capabilities in CG3D.

\section{Conclusion}

In this paper, we proposed a new framework CG3D (CLIP goes 3D), where a 3D Encoder is introduced into the CLIP framework. This 3D Encoder is trained such that the extracted 3D features align with the image and text features of the same category. We also proposed using learnable visual prompts to shift the rendered image distribution to that of the CLIP to get better representative image features from the visual encoder. Through extensive analysis, we demonstrate the zero-shot capabilities of CG3D, which enables zero-shot 3D classification, scene querying with natural language, and cross-modal retrieval. Furthermore, CG3D provides strong initial weights when training 3D networks for downstream tasks.

%%%%%%%%% REFERENCES
{\small
\bibliographystyle{ieee_fullname}
\bibliography{arxiv_v0}
}

\newpage
\newpage

\appendix

\section{Scene querying with language}
We provide further qualitative results that demonstrate the language-based querying capabilities of our proposed framework on point clouds of indoor scenes. Note that in the main paper, we pre-trained on ShapeNet which is not a real-world dataset. In order to imporve performance on scene-querying on real-world datasets, we perform additional pre-training on the ScanObjectNN dataset for querying on a collection of meshed indoor scenes from the S3DIS  \cite{2017arXiv170201105A} and ScanNet \cite{dai2017scannet} datasets.  

% \begin{e*}figure*}
%     \centering
%     \includegraphics[width=\linewidth]{figures/loss_supp.drawio.pdf}
%     \caption{Caption}
%     \label{fig:loss}
% \end{figur
\begin{figure*}[!htp]
    \centering
    \includegraphics[width=0.8\linewidth]{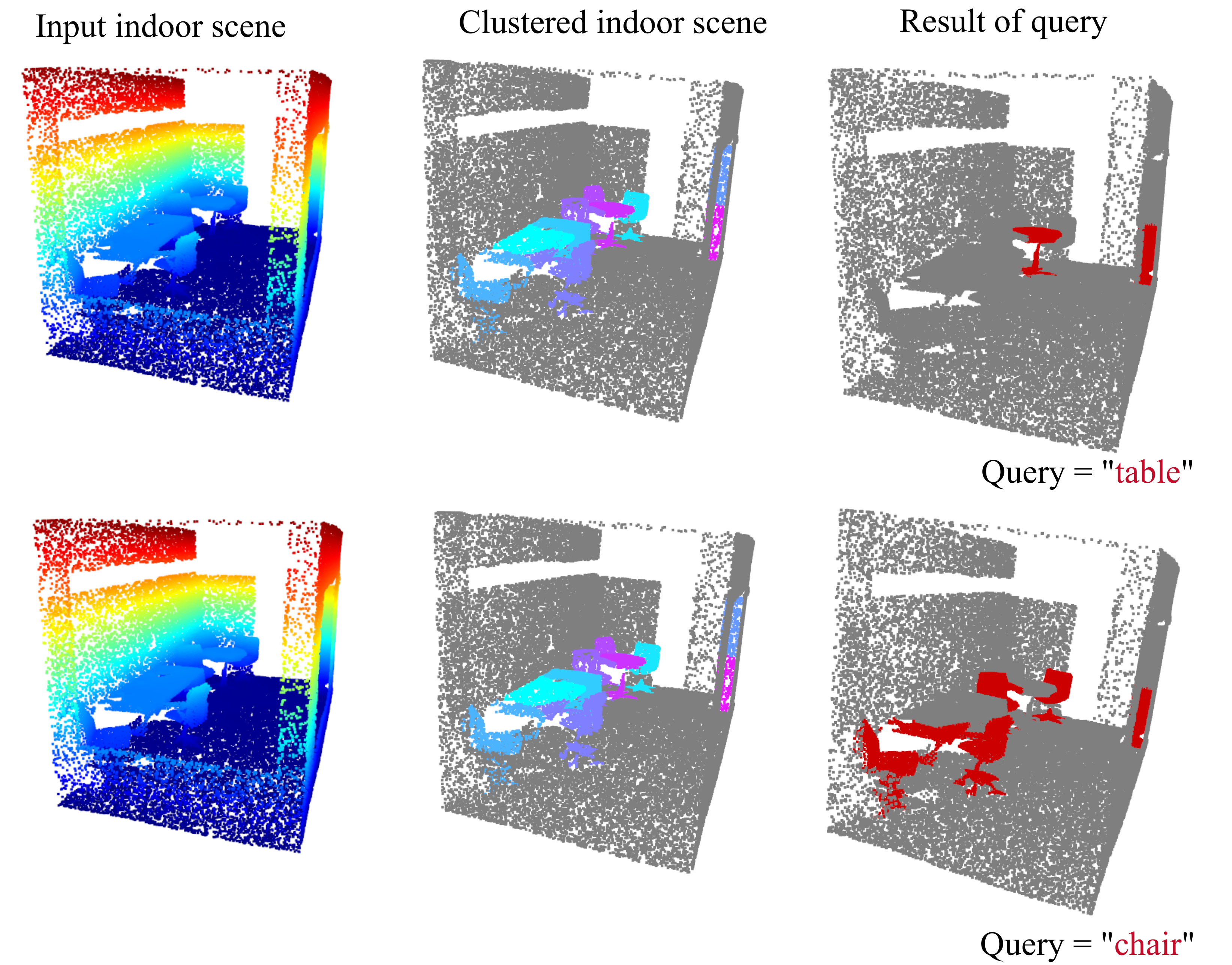}
    \caption{Qualitative results of scene-querying samples from the S3DIS \cite{2017arXiv170201105A} dataset.}
    \label{fig:s3dis}
\end{figure*}
\begin{figure*}
    \centering
    \includegraphics[width=0.8\linewidth]{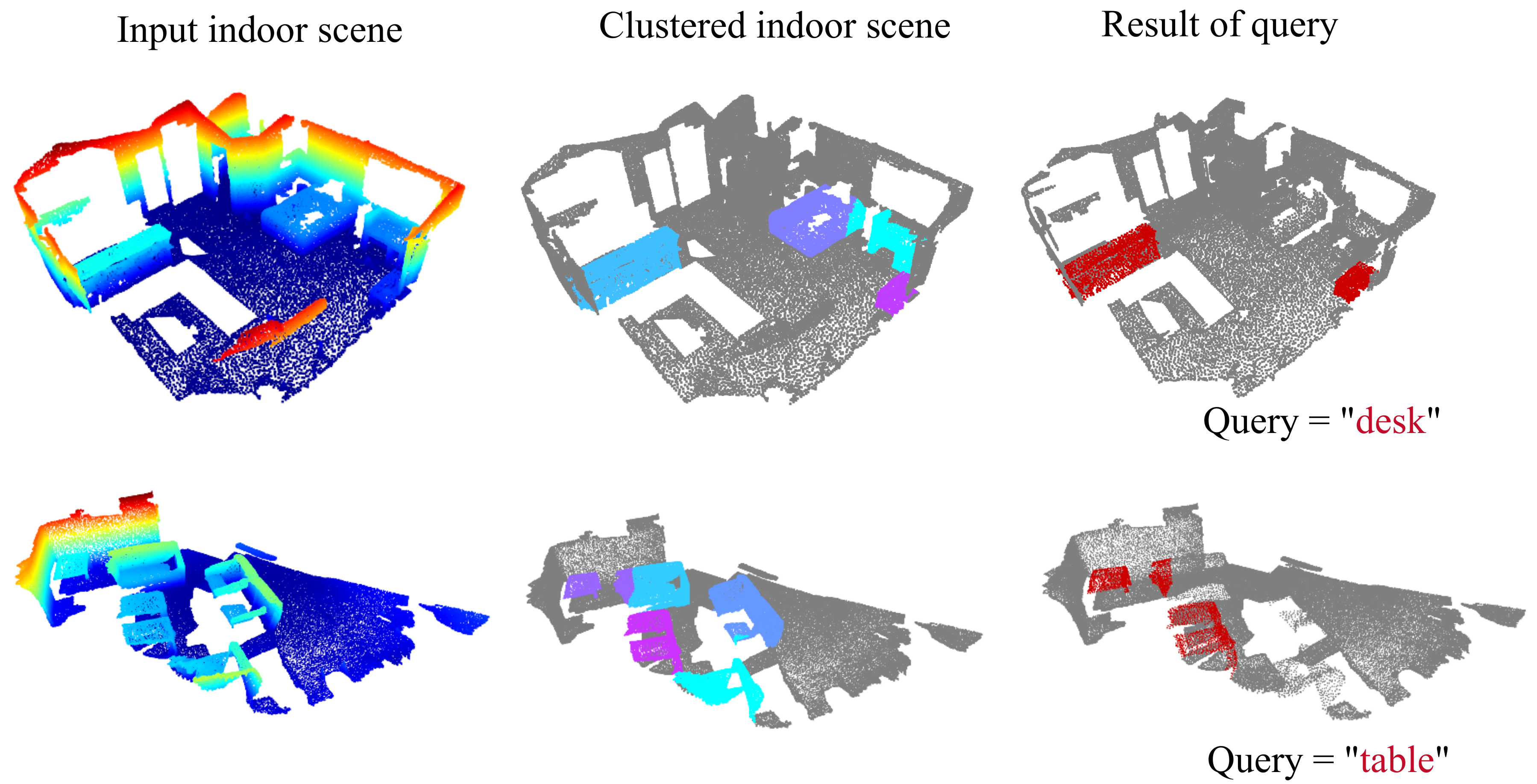}
    \caption{Qualitative results of scene-querying samples from the ScanNet \cite{dai2017scannet} dataset.}
    \label{fig:scan}
\end{figure*}

% \subsection{Pre-training on ScanObjectNN}

During standard CG3D pre-training, we render the textured CAD models of ShapeNet to use as inputs to our visual encoder. Such rendering is not possible with ScanObjectNN, so we project each point cloud to a depth map in a random view. The text caption is curated in the standard procedure.

\subsection{Scene querying on S3DIS}
We perform language-based scene querying on the indoor scene dataset S3DIS \cite{2017arXiv170201105A}. Each scene (disregarding the floor and ceiling regions, as is standard in semantic segmentation tasks) is clustered into regions, each of which is passed to CG3D along with a query containing the object to be localized. Some qualitative results may be seen in Figure \ref{fig:s3dis}. Each row shows an input indoor scene, the result of clustering, and the final result of the language query. We query the same scene for two different objects. In the second row, it can be observed that several instances of the queried category ``chair" are correctly identified. 

\subsection{Scene querying on ScanNet}
We also demonstrate the performance of the 3D encoder pre-trained on real data on indoor scene samples from the ScanNet \cite{dai2017scannet} dataset. Figure \ref{fig:scan} shows the result of querying on two samples. The quality of the results depends on the clustering accuracy, which can cause spurious results. 
 However, both instances of the queried object are correctly identified in both examples.

\section{Leveraging prompt-tuning for images}
% Training CLIP's visual encoder through the tuning of prompt tokens enables improved zero-shot capabilities for images in addition to the established zero-shot capabilities of the 3D encoder. In order to leverage the visual encoder on 3D shapes, we project each point cloud to a 2D depth map. In Table \ref{tab:} we compare the .... [TODO]
In the main paper, we showed how we can train a 3D encoder in the CLIP framework and illustrated its benefits. We also introduced prompt tuning in the 2D encoder of CG3D to tune it towards rendered 3D shapes and objects. Performing contrastive visual prompt tuning not only aids in pre-training the 3D encoder, but also allows us to leverage CLIP's visual encoder for 3D shapes. Using visual prompts allows us to train the visual encoder without forgetting the already existing weights of CLIP. Thus, this helps us obtain a new visual encoder that has a capability of performing better than normal CLIP encoder on image-based 3D tasks. By image-based 3D tasks, we mean applications such as PointCLIP \cite{zhang2021pointclip} where  zero-shot classification is performed by forwarding depth maps of each point cloud directly to the visual encoder.

\begin{table}[htbp]
\resizebox{\columnwidth}{!}{%
\begin{tabular}{cccc}
\hline
\multirow{2}{*}{Method}&\multirow{2}{*}{Backbone} & \multicolumn{2}{c}{Overall Acc} \\ \cline{3-4} 
                &        & ShapeNet \cite{shapenet2015}     & ModelNet40 \cite{wu20153d} (ZS)    \\ \hline
PointCLIP \cite{zhang2021pointclip}   &    ViT-B       &   33.6           &           10.1       \\
CG3D-render  &      ViT-B + prompt     &      77.8        &     24.8             \\
CG3D-depth   &   ViT-B + prompt        &     37.0         &        34.4          \\ \hline

\end{tabular}
}
\caption{Comparison of classification performance of CLIP visual encoder and CG3D visual encoder on the ShapeNet and ModelNet datasets. CG3D is pretrained on ShapeNet.}
\label{tab:img}

\end{table}

\begin{table}[htbp]
\resizebox{\columnwidth}{!}{%
\begin{tabular}{cccc}
\hline
\multirow{2}{*}{Method} & \multirow{2}{*}{Backbone} & \multicolumn{2}{c}{Linear probe acc} \\ \cline{3-4} 
                        &                           & ShapeNet       & ModelNet40       \\ \hline
PointCLIP               & ViT-B                     &      70.31          &      43.31               \\
CG3D-render             & ViT-B + prompt            &    82.47            &   46.15                  \\
CG3D-depth              & ViT-B + prompt            & 70.38          & 68.80               \\ \hline
\end{tabular}
}
\caption{Comparison of linear probe classification performance of CLIP visual encoder and CG3D visual encoder on the ShapeNet and ModelNet datasets.}
\label{tab:lp}

\end{table}

 % Methods that apply CLIP to 3D shapes such as PointCLIP \cite{zhang2021pointclip} perform zero-shot classification by forwarding depth maps of each point cloud directly to the visual encoder. Since CLIP is trained on natural images, there exists a gap in distribution from the test data, which affects performance. By injecting learnable prompt tokens in each transformer layer of the visual encoder, we tune CLIP to operate on images of 3D shapes. We now have a CLIP image encoder specific to 3D shapes along with the 3D encoder.
% \begin{table}[]
% \resizebox{\columnwidth}{!}{%
% \begin{tabular}{cccc}
% \hline
% \multirow{2}{*}{Image backbone} & \multicolumn{3}{c}{ZS acc} \\ \cline{2-4} 
%                                 & MN10 & MN40 & ScanObjectNN \\ \hline
% ViT-B + shallow prompt          &   64.2   &   43.2   &        20.5      \\
% ViT-B + deep prompt             &  67.3    &   50.6   &     25.6         \\ \hline
% \end{tabular}
% }
% \caption{Ablation study on visual prompting.}
% \label{tab:vpt}

% \end{table}

%\multicolumn{1}{c|}{67.3}       & \multicolumn{1}{c|}{50.6}      & \multicolumn{1}{c}{25.6}

We examine the effectiveness of tuning the visual encoder using our method by comparing the classification and zero-shot classification performance on the ShapeNet and ModelNet datasets respectively. In column 3 of Table \ref{tab:img}, we compare the overall classification accuracy of single-view PointCLIP \cite{zhang2021pointclip} against the prompt-tuned visual encoder on rendered images of ShapeNet objects. Column 4 compares zero-shot classfication performance on the 40-class split of ModelNet. CG3D-render denotes that the framework uses rendered images of ShapeNet objects during pre-training. CG3D-depth denotes that depth projections are used as inputs to the visual encoder during pre-training.

We observe that performing visual prompt tuning of the 2D encoder significantly improves the performance of CLIP visual encoder in CG3D for 3D shape datasets. Classification performance on rendered ShapeNet images improves by 44.2\% points and and zero-shot performance on ModelNet depth images improves by 24.3\% points. The best performance is observed when the modality of the pre-training image dataset aligns with the modality during evaluation. Nonetheless, we observe improvements even when the modalities do not match.

As a further demonstration of the quality of image features obtained after visual prompt tuning, we perform linear probing as done in \cite{radford2021learning} by performing logistic regression on the learned image features from the visual encoder. In Table \ref{tab:lp}, we compare the linear probe performance using image features from the out-of-the-box vision encoder of CLIP (as used in \cite{zhang2021pointclip}) with the visual encoder with tuned prompt tokens. A trend similar to Table \ref{tab:img} can be observed, where prompt tuning results in improved classification performance, with the best performance occurring when the modalities of the train and test data match. This means that the visual encoder is able to effectively learn image features specific to 3D shapes. This opens up numerous possibilities in multi-modal learning between 3D shapes and images.

% \section{Ablations on prompt-tuning}
% We examine further the role prompt tokens have on zero-shot performance. We implement two variations of visual prompt tuning on transformers \cite{jia2022visual}, namely shallow and deep prompts. While using shallow prompts, learnable prompt tokens are injected only at the input to the visual encoder. In deep prompting, a series of learnable tokens are input at the input of each encoder in the transformer block. Table \ref{tab:vpt} shows a comparison of zero-shot performance of the PointTransformer 3D encoder when trained under the CG3D framework using each prompting strategy. The use of deep prompts clearly aids 3D zero shot performance, particularly in the case of ScanObjectNN \cite{scanobjectnn}, where there is a 5.1\% point increase from shallow prompting. This means that more learnable parameters at every stage of the transformer encoder results in a visual encoder that provides more meaningful image features for contrastive training with shape features. This observation leads us to choose deep prompting as the design for tuning the visual encoder. 

\end{document}